\documentclass{article}

\usepackage{spconf,amsmath,graphicx}
\usepackage{amsfonts}
\usepackage{bm, bbm}
\usepackage{color}
\usepackage{booktabs}
\usepackage{array}
\usepackage[caption=false, font=footnotesize]{subfig}
\usepackage{dblfloatfix}
\usepackage{url}
\usepackage{textcomp}
\usepackage{multirow}
\usepackage{cite}
\usepackage{theorem}

\newtheorem{theorem}{Theorem}[section]
\newtheorem{definition}[theorem]{Definition}
\newtheorem{proposition}[theorem]{Proposition}

\newcommand{\RR}{\ensuremath{\mathbb R}}

\newcommand{\Tr}{\ensuremath{\operatorname{Tr}}}

\newcommand{\corr}[1]{\textcolor{black}{#1}}

\let\OLDthebibliography\thebibliography
\renewcommand\thebibliography[1]{
  \OLDthebibliography{#1}
  \setlength{\parskip}{0pt}
  \setlength{\itemsep}{0pt plus 0.3ex}
}

\begin{document}\sloppy

\def\x{{\mathbf x}}
\def\L{{\cal L}}

\title{Multilayer Graph Clustering with Optimized Node Embedding}

%
\name{Mireille El Gheche and Pascal Frossard}
\address{Ecole Polytechnique F\'ed\'erale de Lausanne (EPFL), LTS4, Lausanne, Switzerland}

\maketitle

\begin{abstract}
	
	We are interested in multilayer graph clustering, which aims at dividing the graph nodes into categories or communities.
	To do so, we propose to learn a clustering-friendly embedding of the graph nodes
	by solving an optimization problem that involves a fidelity term to the layers of a given multilayer graph, and a regularization on the (single-layer) graph induced by the embedding.
	The fidelity term uses the contrastive loss to properly aggregate the observed layers into a representative embedding. The regularization pushes for a sparse and community-aware graph, and it is based on a measure of graph sparsification called ``effective resistance'', coupled with a penalization of the first few eigenvalues of the representative graph Laplacian matrix to favor the formation of communities. The proposed optimization problem is nonconvex but fully differentiable, and thus can be solved via the descent gradient method. Experiments show that our method leads to a significant improvement w.r.t. state-of-the-art multilayer graph clustering algorithms.
	
\end{abstract}

\begin{keywords}
	Multilayer graph, embeddings, clustering, contrastive loss, $K$-components, effective resistance.
\end{keywords}

\section{Introduction}
\label{sec:intro}

Graph\corr{s} are getting increasingly popular in computer science and related fields. They have emerged as a fundamental representation and an essential tool to understand, describe, model and analyse the complex system of interacting entities that arises in sociology, biology, recommendation systems, economics, to cite a few examples. In all of these domains, data points are represented as vertices, and links between the nodes are represented as edges. 

Many real-world applications often rely on multiple sources of information, leading to multiple graphs that describe different types of relationship between the same set of nodes. 
This structure is hereby referred to as multilayer graph. For example, a social network
system can be represented by a multilayer graph, where
the nodes are the users, and each layer describes a different
kind of relationship (geographic distance, relational information,
behavioral relationships based on user actions or interests,
etc). A public transport system can be also represented by a multilayer graph, where the nodes are transportation hubs and each layer describes a different mean of transportation (a bus
line, a metro line, etc.).

Multilayer graphs are not only useful as data representation forms, they also play a crucial role in many machine learning and data mining tasks. The main challenge in machine learning on graphs is finding a way to integrate information about the graph into a machine learning model. The traditional way to extract information from graphs is by computing graph statistics (e.g., degrees, connectives), kernel functions, or handcraft features to measure local structures. 

More recently, there has been a surge of approaches that seek to learn graph \textit{embeddings} for representing the graph nodes into a low-dimensional space. The majority of the methods for multilayer graph clustering hinges around low-rank matrix factorization \cite{Shen2018}, probabilistic generative models \cite{NIPS2012_0a1bf96b} or interpretation of Principal Component Analysis (PCA) on graphs \cite{Saerens2004}, which links the graph structure to a subspace spanned by the top eigenvectors of the graph Laplacian matrix. Moreover, numerous methods have been proposed in the literature for representation learning on graphs, such as random walks \cite{Perozzi2014} and deep learning approaches \cite{node2vec2016, Pfau2019}. 
One of the most popular approaches is spectral clustering \cite{elgheche2019_implicit}, which consists of embedding the graph vertices into a subspace spanned by the eigenvectors of the graph Laplacian matrix corresponding to the $K$ smallest eigenvalues, where clusters can be easily detected via the $K$-means algorithm \cite{Macqueen1967}. The spectral clustering methods rely on the graph Laplacian matrix, and thus their performance highly depends on the quality of this representation, for which many variants have been proposed. 

The straightfoward way to extend spectral clustering to multilayer graphs is to linearly combine  different layers into a single representative layer \cite{Chen_Hero_2017}. However, this method may not be able to properly capture the structure information present in each layer. In order to properly take into account the topology shared across layers, one can see the graph layers as points of a Grassman manifold \cite{Wang_TIP_2013} or as points in a symmetric positive definite manifold \cite{TSIPN_elgheche2019}. The learning framework in \cite{Wang_TIP_2013,TSIPN_elgheche2019} can be formulated as the joint optimization problem of finding the graph that is representative of all layers, and the embedding that allows for the clustering of its nodes. Nevertheless, this task is complex to solve.



The work in \cite{gheche2020multilayer} presents a general approach for learning a community-based representative graph of a given multilayer graph. It works in two steps: aggregation of the Laplacian matrices of the multilayer graph, and then performing the spectral clustering of the resulting (single-layer) graph. To do so, the authors have combined three ideas: graph learning, graph sparsification and node clustering. The resulting optimization model is well adapted to graph learning with sparse and community structure. 
Differently from \cite{gheche2020multilayer}, we present a general approach for learning a multilayer graph embedding, while integrating the information of multiple views to offer an effective way to cluster nodes of a multilayer graph. In particular, we formulate the clustering problem as an optimization problem involving a contrastive loss as a data fidelity term w.r.t.\ the observed layers, and two problem-specific regularization terms on the (single-layer) graph induced by the embedding.
The resulting optimization problem is efficiently solved via gradient descent. Experimental results show that the proposed approach achieves a better clustering performance compared to baseline multilayer graph clustering approaches, due to the effective combination of the different layers into a single graph capturing the clusters that would otherwise be hidden if each layer was considered alone. 

The paper is organized as follows. Section \ref{sec:problem_formulation} details the problem formulation for learning  graph embeddings given a multilayer graph. Section \ref{sec:experiments} provides an experimental validation of the proposed approach on synthetic and real multilayer graphs. Section \ref{sec:conclusion} draws the conclusion.

\section{Preliminaries}
\label{sec:problem_formulation}

\label{subsec:preliminaries}

%
\paragraph*{Notation}

Let $\mathcal{G}$ be a multilayer graph with a set $V$ of $N$ vertices shared across $S \geq 1$ layers of edges, denoted as
\begin{equation}
\mathcal{G} = \big\{\mathcal{G}^s(V,E^s)\big\}_{1\le s\le S}.
\end{equation} 
For each layer $s\in\{1,\dots,S\}$, $\mathcal{G}^s$ is an undirected weighted graph with the shared vertex set $V$, a separate edge set $E^s$, and an adjacency matrix $W^s = [w^s_{i,j}] \in \RR^{N\times N}$ whose element $w^s_{i,j}$ is positive if there is an edge between vertex $i$ and $j$, and zero otherwise. The weighted degree of a vertex $i$ is defined as the sum of the edge weights incident to $i$ in the graph $\mathcal{G}^s$, i.e., $d^s_i = \sum_j w^s_{ij}$. The degree matrix $D^s$ is then defined as
\begin{equation}
D^s_{i,j} = \begin{cases} d^s_i \quad & \textup{if $i=j$} \\
0 \quad & \textup{otherwise}
\end{cases}
\end{equation}
and the Laplacian matrix of $\mathcal{G}^s$ is defined as 
\begin{equation}
L^s=D^s-W^s.
\end{equation}


%
\paragraph*{Node Embedding}

The purpose of graph node embedding is to encode graph nodes as low-dimensional vectors that summarize the structure of their local graph neighborhood. Specifically, the embedding is a matrix containing the representative vectors for all the graph nodes 
$$
\textbf{Z} = \begin{bmatrix}
\textbf{z}_1^\top\\
\vdots\\
\textbf{z}_N
\end{bmatrix}
\in \RR^{N\times K}.
$$ 

In our setting, we require that the embedding provides a clustering-friendly representation of the graph nodes, in the sense that the above vectors should be trivially clustered with a standard algorithm such as $K$-means. Ideally, we would like the embedding to be only populated by indicator vectors, with the $n$-th row of $\textbf{Z}$ having the value $1$ in the $k$-th position (and $0$ otherwise) if the node $n$ belongs to the cluster $k$. This is however a very stringent requirement, that we relax by only requiring the embedding to be a semi-orthogonal matrix
$$
\textbf{Z}^\top \textbf{Z} = \frac{1}{N} I_{K\times K}.
$$


Generally speaking, the geometry of a low-dimensional embedding relates to the \textit{edge similarity} of the different layers, in the sense that the similarity between two vectors $\textbf{z}_i$ and $\textbf{z}_j$ must be proportional to the connection strength observed in the edges between the nodes $i$ and $j$ among the different layers.
The simplest way to measure the similarity between two vectors is to use the inverse Euclian distance \cite{NIPS2001_f106b7f9}
\begin{equation}
\label{eq:l2norm}
\textrm{SIM} (\textbf{z}_i, \textbf{z}_j) = -\|\textbf{z}_i - \textbf{z}_j\|^2,
\end{equation}
or the inner product \cite{kruskal1964}
\begin{equation}
\label{eq:scalar_product}
\textrm{SIM} (\textbf{z}_i, \textbf{z}_j) = \textbf{z}_i^\top \textbf{z}_j.
\end{equation}
A different approach consists of 
using a statistical-oriented measure of similarity \cite{node2vec2016, Goyal_2018}
\begin{equation}
\textrm{SIM} (\textbf{z}_i, \textbf{z}_j) = \frac{\exp^{\textbf{z}_i^\top \textbf{z}_j}}{\sum_{k\in\mathcal{N}_i} \exp^{\textbf{z}_i^\top \textbf{z}_k}}.
\end{equation}
Another successful approach was introduced in \cite{line_tang_2015}, where the similarity measure is based on the sigmoid function
\begin{equation}
\textrm{SIM} (\textbf{z}_i, \textbf{z}_j) = \frac{1}{1 + \exp^{-\textbf{z}_i^\top \textbf{z}_j}}.
\end{equation}
The sigmoid-based similarity has the nice feature of being bounded in the interval $]0,1[$.  
In the following, we will define a conceptually analogous measure for our approach. 


\section{Multilayer Clustered Node Representations}
\label{subsec:learning_clustered_node_emb}

The core operation for learning a clustered node representation of a multilayer graph is to embed the vertices into a low-dimensional space, where the projected points can be trivially clustered. More specifically, assuming that $K$ is the number of desired clusters, the goal is to find an embedding matrix $\textbf{\textbf{Z}}\in\mathbb{R}^{N\times K}$ such that each row $\textbf{z}_n \in \mathbb{R}^K$ is a vector representing a graph vertex. By doing so, the $K$-mean algorithm can be then applied on the rows of $\textbf{Z}$, so that $K$ clusters are formed by grouping together the vertices that are strongly connected in the given multilayer graph. 

We address the problem of analyzing multilayer graphs by proposing a method for learning
a set of node embeddings $\textbf{Z}$ that allows us to identify communities or clusters in the multilayer graph.
Specifically, we estimate the graph embedding $\textbf{Z}$ from the multilayer graph $\mathcal{G}$ by solving the following optimization problem
\begin{equation}
\label{eq:laplacian_learning}
\operatorname*{minimize}_{\textbf{Z}\in\RR^{N\times K}}\; \sum_{s=1}^S {J}(\textbf{Z};\mathcal{G}^s) + \mathcal{R}(\textbf{Z})
\quad{\rm s.t.}\quad
\textbf{Z}^\top \textbf{Z} =  \frac{1}{N} I_{K\times K}.
\end{equation}
where ${J}(\cdot\,; \mathcal{G}_s)$ is a data-fidelity term 
with respect to layer $s$, and $\mathcal{R}$ is a suitable regularization term. We use the contrastive loss as a data fidelity term, in order to properly aggregate the observed layer information into a representative embedding. The regularization is based on a measure of graph sparsification called ”effective resistance”, coupled with a penalization to favor the formation of communities.

\subsubsection{Data fidelity term}

In order to properly represent the observed layers as a set of node embeddings, we employ the contrastive loss. It intends to maximize the likelihood of preserving the node neighborhoods observed in each graph layer.
The contrastive loss was originally introduced in \cite{node2vec2016}, and used later for self-supervised learning \cite{khosla2020supervised, Chen2020ASF, He_2020_CVPR}. Let $\mathcal{N}^s(i)$ be the set of nodes that appear in the neighboring of a node $i$. The contrastive loss for graph node embedding reads
\begin{equation}
\label{eq:contrastive_emb}
\mathcal{J}(\textbf{Z}; \mathcal{G}^s ) = \sum_{i=1}^N \sum_{j \in \mathcal{N}^s(i)} -\log\left( \dfrac{\exp\big( \textrm{SIM}(\textbf{z}_i,\textbf{z}_j)\big)}{\sum_{k\neq i}\exp\big(\textrm{SIM}(\textbf{z}_i,\textbf{z}_k)\big)} \right),
\end{equation}
where the measure to quantify the similiarity of any two node embedding $\textbf{z}_i$and $\textbf{z}_j$ is given by
\begin{equation}\label{eq:proposed_similarity}
\textrm{SIM} (\textbf{z}_i, \textbf{z}_j) = \frac{1}{1 + \exp^{\|z_i-z_j\|^2}}.
\end{equation}
The contrastive loss aims at learning the graph node embeddings in such a way that the neighbors in the multilayer graph are pulled together and non-neighbors are pushed apart.

\subsubsection{Regularization}

Instead of imposing a regularization on the embedding, we propose to introduce a regularization on the graph induced by the embedding. This allows us to maintain a structural coherence with the fidelity term, which is defined on a multilayer graph rather than some data points. Another benefit of this approach is that the final embedding induces a meaningful single-layer graph summarizing a given multilayer graph.

A possible way to build a graph 
from the embedding $\textbf{Z}$ consists of computing the pairwise distance between the embedding vectors such that 
\begin{equation}
\corr{\mathcal{Z}_{i,j} = \textrm{SIM}(\textbf{z}_i,\textbf{z}_j).}
\end{equation}
Dealing with the matrix $\mathcal{Z} \in \RR^{N\times N}$, in order to incorporate some properties on the embedding space, is challenging and not straightforward. Hence, one can convert $\mathcal{Z}$ to the space of valid Laplacian matrices $\mathcal{C}$\footnote{The set of valid combinatorial graph Laplacian matrices defined as
	\begin{equation}
	\mathcal{C} = \{L\in\mathbb{R}^{N\times N} \,|\, \; L_{ij}=L_{ji}\leq 0, L_{ii}=-\sum_{j\neq i} L_{ij}\}.
	\end{equation}}
\begin{equation}
\label{eq:graph_learning}
L(\textbf{Z}) = \mathcal{L}(\rm{triu} \mathcal{Z})
\end{equation}
where $\mathcal{L}\colon \RR^{(N-1)N/2}\rightarrow \RR^{N\times N}$ is the linear operator that converts the upper-triangular part of a matrix into the corresponding Laplacian matrix. 

We aim at analyzing multilayer graphs by properly combining the information provided by the embedding $\textbf{Z}$, while preserving the specific structure that allows us to eventually identify communities or clusters that are crucial in the analysis of graph data. To do so, we use $L(\textbf{Z})$ and two pieces of information: the effective resistance $\mathcal{R}_{\rm eff}$ to measure the sparsification of a community regularization $\mathcal{R}_{\rm com}$ to favor the formation of communities:
\begin{equation*}
\mathcal{R}(\textbf{Z}) = \gamma_{1} \mathcal{R}_{\rm eff}(L(\textbf{Z})) + \gamma_{2} \mathcal{R}_{\rm com}(L(\textbf{Z}))
\end{equation*}
Our first regularization is an interesting graph measure derived from the field of electric circuit analysis, where it is defined as the accumulated effective resistance between all pairs of vertices \cite{friedman_sparse_2008,Tarzanagh2017}. Given a graph Laplacian matrix $L\in\RR^{N\times N}$, the effective resistance $R_{ij}$ between a pair of nodes $i$ and $j$ is defined as
\begin{equation}
R_{ij} = v_i - v_j,
\end{equation}
where the vector $v=[\dots, v_i, \dots, v_j, \dots]^\top\in\RR^{N}$ is the solution to the equation
\begin{equation}\label{eq:vsol}
Lv = \delta_i - \delta_j,
\end{equation}
with $\delta_n$ being the $n$-th column of the $N\times N$ identity matrix. Note that all solutions to Eq.\ \eqref{eq:vsol} give the same value of $v_i - v_j$. Now, let us define the pseudoinverse of $L$ as
\begin{equation}
L^\dagger = (L + \mathbbm{1}\mathbbm{1}^\top/N)^{-1} - \mathbbm{1}\mathbbm{1}^\top/N,
\end{equation}
where $\mathbbm{1}=[1,\dots,1]^\top \in \RR^N$. Then, the effective resistance can be rewritten as \cite{klein1993}
\begin{equation}
R_{ij} = (\delta_i - \delta_j)^\top L^\dagger (\delta_i - \delta_j),
\end{equation}
and the total effective resistance amounts to \cite{ghosh2008}
\begin{equation}
R_{\rm tot} = \sum_{i < j} R_{ij} = N \Tr(L^\dagger) = N \sum_{n=2}^N \frac{1}{\lambda_n},
\end{equation}
where $0=\lambda_1<\lambda_2\le\dots\le\lambda_N$ are the eigenvalues of $L$. In this paper, we adopt a regularization that is proportional to the total effective resistance, namely
\begin{equation}
\mathcal{R}_{\rm eff}(L) = \sum_{n=K+1}^N \frac{1}{\lambda_n(L)},
\end{equation}
where $\lambda_n(L)$ denotes the $n$-th eigenvalue of $L$, and $K\ge1$ is equal to the number of communities chosen in the regularization discussed next.

While graph sparsification is widely used, especially in high-dimensional settings, it is not enough to learn a graph with a specific structure. Hence, our  second regularization favors the formation of communities by studying the eigenvalues of the Laplacian matrix. 
Spectral graph theory is one of the main tools to study the relationship between the eigenvalues and the structure of a graph. We will now recall two basic facts of spectral graph theory. Let $L$ be the Laplacian matrix of an undirected graph, along with the corresponding eigenvectors  $U=[u_1, \cdots, u_N]$ and eigenvalues $\Lambda= {\rm Diag}(\lambda_1, \cdots, \lambda_{N})$. 
\begin{definition}
	A  connected  component  of  an  undirected  graph  is  a  connected subgraph such that there are no edges between vertices of the subgraph and vertices of the rest of the graph.
\end{definition}

\begin{proposition}
	A graph has $K$ connected components if its vertex set can be partitioned
	into $K$ disjoint subsets such that any two nodes belonging to different subsets are not connected. The eigenvalues of its Laplacian matrix are then as follows
	\begin{equation}
	0 = \lambda_1 = \dots = \lambda_K \qquad 0 < \lambda_{K+1} \le \dots \le \lambda_N.
	\end{equation}
\end{proposition}
To favor the learning of a graph with $K$ communities that are not necessarily disconnected, we propose to minimize the first $K$ eigenvalues of the corresponding Laplacian matrix, yielding the following regularization
\begin{equation}
\mathcal{R}_{\rm com}(L) = \sum_{n=1}^K \lambda_n^2(L),
\end{equation}
where $K\ge 1$, and $\lambda_n(L)$ denotes the $n$-th eigenvalue of $L$. 


\section{Experiments}
\label{sec:experiments}

\begin{figure}[!h]
	\centering
	\subfloat[Multilayer graph with $S=3$ layers. From the \textit{left} to the \textit{right}: layer 1, layer 2, and layer 3. The colors indicate the groundtruth clusters. \label{fig:gaussian_layers}]{\includegraphics[width=0.8\linewidth]{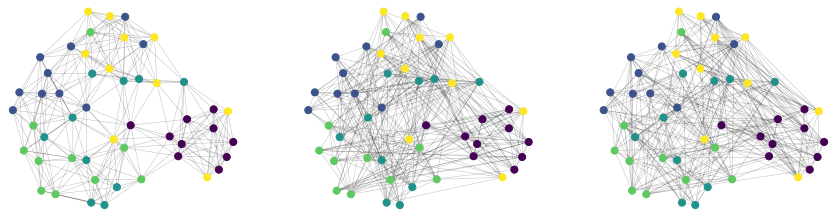}}
	\hfill
	\subfloat[Arithmetic mean]{\includegraphics[width=0.48\linewidth]{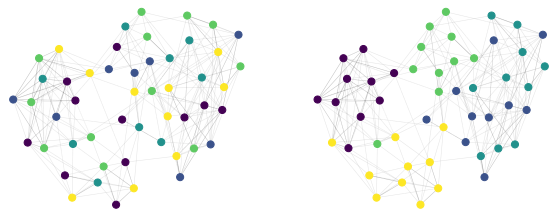}}
	\hfill
	\subfloat[Projection mean \cite{Dong_TSP_2014}]{\includegraphics[width=0.48\linewidth]{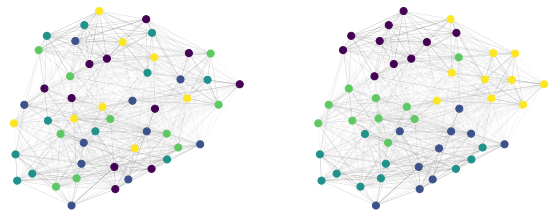}}
	\hfill
	\subfloat[Geometric mean \cite{TSIPN_elgheche2019}]{\includegraphics[width=0.48\linewidth]{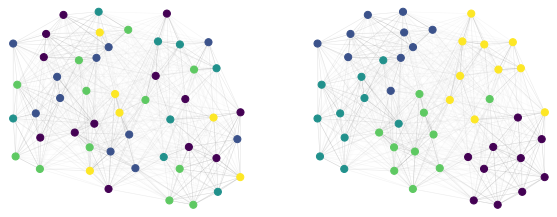}}
	\hfill
	\subfloat[Clustered Multilayer \cite{gheche2020multilayer}]{ \includegraphics[width=0.48\linewidth]{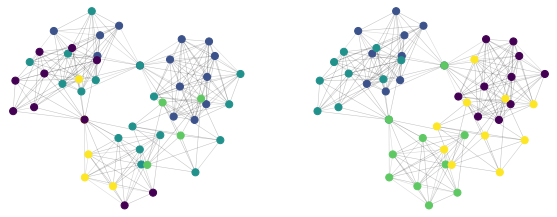}}
	\hfill
	\subfloat[Proposed approach.]{\includegraphics[width=0.48\linewidth]{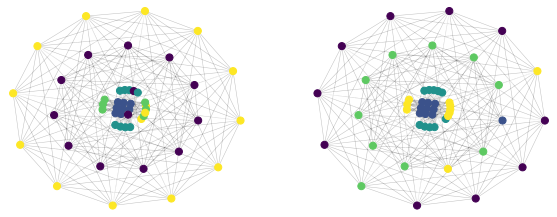}}
	
	\caption{Illustrative example of multilayer representative graph learning. (a): the original multilayer graph composed of three layers. (b)-(c)-(d)-(e)-(f): \textit{left} the clustering results with $K=5$ on each of the estimated representative graph, \textit{right} the groundtruth labels. Numerical evaluations are reported in Table \ref{tab:synt}.}
	\label{fig:gaussian}
\end{figure}


\subsection{Settings}
We compare our approach with four clustering algorithms. The methods referred to as arithmetic mean (average), the projection mean \cite{Dong_TSP_2014}, the geometric mean \cite{TSIPN_elgheche2019}, and clustered multilayer \cite{gheche2020multilayer} 
work in two steps: they aggregate the Laplacian matrices of the multilayer graph, and then perform the spectral
clustering of the resulting (single-layer) graph. The difference
lies in the aggregation step, which is performed in Grassman
manifold\cite{Dong_TSP_2014}, in the SPD manifold \cite{TSIPN_elgheche2019}, or using some regularizations on the node clustering and graph sparsification as in clustered multilayer method \cite{gheche2020multilayer}. All the methods are used with their default hyperparameters. As for our method\footnote{The parameters $\gamma_1$ and $\gamma_2$ are hand-tuned so as to obtain the best numerical results.}, we implement the algorithm with $\gamma_1 = 0.1$ and $\gamma_2=100$. The orthogonality constraint in Problem \eqref{eq:laplacian_learning} can be enforced implicitly, leading to a differentiable cost function that can be optimized via gradient descent \cite{TSIPN_elgheche2019}. 
We use four criteria to measure the clustering performance: Accuracy, Purity, Normalized Mutual Information (NMI), and adjusted Rand Index (RI). They measure the agreement of two partitions, ignoring permutations and with no requirement to have the same number of clusters. Values close to zero indicate two assignments that are largely independent, while values close to one indicate significant agreement. All the experiments are conducted in Python/Numpy/PyTorch on a 40-core Intel Xeon CPU at 2.5 GHz with 128GB of RAM.

\subsection{Datasets}

In our experiments, we consider one synthetic dataset. 
\begin{itemize}
	\item \textit{Synthetic dataset}. It consists of $S=3$ point clouds of arbitrary size $N=50$, each generated from $2$-dimensional Gaussian mixture model with $K=5$ components having different means and covariance matrices. We build a $20$-nearest neighbor ($K$-NN) graph on each point cloud and we set the edge weights to the reciprocal of the Euclidean distance between pairs of neighbors. This gives us $3$ layers as shown in Figure \eqref{fig:gaussian_layers}.
\end{itemize}
Then we consider two real datasets. The Newsgroup  \textit{NGs}\footnote{http://lig-membres.imag.fr/grimal/data.html} and the  \textit{BBC sport} \cite{bbcsport_2009} datasets. 
\begin{itemize}
	\item \textit{NGs dataset}. The newsgroup dataset is a subset of the 20 Newsgroup dataset. \textit{NGs} consists of a collection of 500 documents. Each raw document was pre-processed based on the text segments (giving three layers). Each layer is partitioned  across 5 different newsgroups, each corresponding to a different topic. 
	\item \textit{BBC sport dataset}. It was collected from BBC Sport website. BBC sport dataset consists of 544 documents. Each document is organized in five sport categories and collected from 2004 to 2005. To build a multilayer graph, we connect the document based on the text segments, leading to $S=2$ graph layers.
\end{itemize}

\subsection{Performance assessment}

Figure \ref{fig:gaussian_layers} gives an illustrative of the synthetic dataset. We present the representative graph computed with five different methods: arithmetic mean, geometric mean, projection mean, clustered multilayer and the proposed approach. The colors of the nodes are the the clustering results with $K=5$ on each of the estimated representative graph. For the sake of evaluation, the groundtruth labeling are shown on each sub-graph. As we can notice, the proposed approach is able to provide a structured graph with rich information between the clusters. The clusters in the representative graph computed with the proposed approach are better than the clusters in the arithmetic mean, geometric mean, and projection mean (Table \ref{tab:synt}). The clustered multilayer method and the proposed approach are the best performers. This is due to the optimization approach used in both methods. Furthermore, 
the proposed approach can take advantage of the information carried out by the embedding vectors and the representative graph, leading to better clustering performance.

Table \ref{tab:synt}, Table \ref{tab:NGs} and Table \ref{tab:bbc_sport} report the numerical comparison with the state-of-art methods. Our proposed approach is markedly better than all the baselines on three datasets, except the \textit{BBC sport} in terms of NMI. On the synthetic dataset, the proposed algorithm is the best performer, whereas the aggregation-based techniques are practically equivalent. This may be related to the fact that the embedding estimated with the proposed approach is more relevant for this kind of data compared with the aggregation followed by spectral clustering. 


\begin{table}[t]
	{\footnotesize
	\centering
	\caption{\textit{Synthetic data} ($N=50$, $S=3$, $K=5$). \label{tab:synt}}
	\begin{tabular}{ l   c  cc   c   c}
		\toprule
		Method & Accuracy & Purity & NMI & RI \\
		\midrule	 
		Arithmetic mean  &  0.32 & 0.36 & 0.11 & 0.0 \\
		Geometric mean \cite{TSIPN_elgheche2019} &  0.32 & 0.34 & 0.11 & 0.0 \\
		Projection mean  \cite{Dong_TSP_2014} & 0.36 & 0.38 & 0.10 & 0.0 \\
		Clustered Multilayer \cite{gheche2020multilayer} & 0.54 & 0.54 & 0.43 & 0.22 \\
		Proposed & 0.82 & 0.82 & 0.68 & 0.56 \\
		\bottomrule	
	\end{tabular}}
	
	{\footnotesize
	\centering
	\caption{\textit{NGs } ($N=500$, $S=3$, $K=5$). \label{tab:NGs}}
	\begin{tabular}{ l   c  cc   c   c}
		\toprule
		Method & Accuracy & Purity & NMI & RI \\
		\midrule	 
		Arithmetic mean  &  0.89 & 0.89 & 0.75 & 0.76 \\
		Geometric mean \cite{TSIPN_elgheche2019} & 0.91 &  0.91 &  0.77 &  0.79 \\
		Projection mean  \cite{Dong_TSP_2014} & 0.89 & 0.89 & 0.73 & 0.75 \\
		Clustered Multilayer \cite{gheche2020multilayer} & 0.62 & 0.68 & 0.61 & 0.41 \\
		Proposed & 0.92 & 0.92 & 0.79 & 0.80 \\
		\bottomrule	
	\end{tabular}}
	\newline
	
	{\footnotesize
	\centering
	\caption{\textit{BBC sport }($N=544$, $S=2$, $K=5$). \label{tab:bbc_sport}}
	\begin{tabular}{ l   c  cc   c   c}
		\toprule
		Method & Accuracy & Purity & NMI & RI \\
		\midrule	 
		Arithmetic mean  &  0.45 & 0.47 & 0.27 & 0.074 \\
		Geometric mean \cite{TSIPN_elgheche2019} & 0.47 & 0.49 & 0.29 & 0.08 \\
		Projection mean  \cite{Dong_TSP_2014} & 0.48 & 0.51 & 0.30 &  0.09\\
		Clustered Multilayer \cite{gheche2020multilayer} & 0.35 & 0.36 &  0.05 &  0.05\\
		Proposed & 0.50 & 0.53 & 0.27 &  0.20 \\
		\bottomrule	
	\end{tabular}}
\end{table}



\section{Conclusion}
\label{sec:conclusion}

In this work, we have proposed an optimization approach in order to estimate a community-based multilayer graph embedding. To do so, we have used an objective function that involves a data fidelity term to the observed layers and a regularization on a graph induced by the embedding pushing for a sparse and community-aware graph. The resulting optimization approach is well adapted to clustered embedding estimation and structured graph learning. Experimental results show a better clustering performance of this approach on diverse datasets compared to state-of-the-art multilayer graph clustering.



\section{Acknowledgment}

We would like to thank Dr. Hemant Tyagi and Dr. Ernesto Araya-Valdivia for the useful discussions on graph embedding. This work was partly supported by Huawei.

\bibliographystyle{IEEEbib}
\bibliography{biblio,biblio2}

\end{document}